\pdfoutput=1

\documentclass[11pt]{article}

\usepackage{insights_ws2021/emnlp2021}

\usepackage{graphicx}
\usepackage{booktabs}
\usepackage{subfig}
\usepackage[textsize=large]{todonotes}
\usepackage{booktabs}
\usepackage{adjustbox}
\usepackage{algorithm}
\usepackage{algpseudocode}
\usepackage[inline]{enumitem}
\usepackage{amsmath,amssymb}
\usepackage[font=small,skip=5pt]{caption}
\usepackage{tabularx} 
\usepackage{longtable}
\usepackage{url}
\usepackage{cleveref}
\usepackage{enumitem}
\usepackage[compact]{titlesec}
\usepackage{multirow}
\usepackage{wasysym}
\usepackage{wrapfig}
\usepackage{xspace}

\usepackage[T1]{fontenc}

\usepackage[utf8]{inputenc}

\usepackage{microtype}

\newcommand{\R}{\ensuremath{\mathbb{R}}}
\newcommand{\removed}[1]{}
\newcommand{\REMOVED}[1]{}
\newcommand{\cbow}{CBOW\xspace}
\newcommand{\sg}{SG\xspace}
\newcommand{\dlurl}{\url{https://github.com/bloomberg/koan}\xspace}
\newcommand{\gensim}{Gensim\xspace}
\newcommand{\mikolov}{\texttt{word2vec.c}\xspace}
\newcommand{\us}{\texttt{kōan}\xspace}

\title{Corrected \cbow Performs as well as Skip-gram}

\author{Ozan \.Irsoy \hspace{0.7cm} Adrian Benton \\
  Bloomberg \\
  \texttt{\{oirsoy,abenton10\}@bloomberg.net} \\ \And
  Karl Stratos \\
  Rutgers University \\
  \texttt{karl.stratos@rutgers.edu} \\}

\begin{document}
\maketitle


\begin{abstract}
    \citet{mikolov2013efficient} observed that
    continuous bag-of-words (CBOW) word embeddings tend to underperform Skip-gram (SG) embeddings, and this finding has been reported in subsequent works.  We find that these observations are driven not by fundamental differences in their training objectives, but more likely on faulty negative sampling CBOW implementations in popular libraries such as the official implementation, word2vec.c, and \gensim.  We show that after correcting a bug in the CBOW gradient update, one can learn CBOW word embeddings that are fully competitive with SG on various intrinsic and extrinsic tasks, while being many times faster to train.
\end{abstract}

\section{Introduction}
\label{sec:introduction}

Pre-trained word embeddings are a standard way to boost performance on many tasks of interest such as named-entity recognition (NER), where contextual word embeddings such as BERT \citep{devlin2019bert} are computationally expensive and may only yield marginal gains.
Word2vec \citep{mikolov2013efficient} is a popular method for learning word embeddings because of its scalability and robust performance. 
There are two main Word2vec training objectives: (1) continuous bag-of-words (\cbow) that predicts the center word by averaging context embeddings, and (2) Skip-gram (\sg) that predicts context words from the center word.\footnote{In this work, we always use the negative sampling formulations of Word2vec objectives which are consistently more efficient and effective than the hierarchical softmax formulations.}

It has been frequently observed 
that while \cbow is fast to train, it lags behind \sg in performance.\footnote{\sg requires sampling negative examples from every word in context, while \cbow requires sampling negative examples only for the target word.} This observation is made by the inventors of Word2vec themselves \citep{word2vec}, and also independently in subsequent works \citep{pennington2014glove,stratos2015model}.
This result is surprising since the \cbow and \sg objectives lead to very similar weight updates. 
This is also contrary to the enormous success of contextual word embeddings
based on masked language modeling (MLM),
as \cbow follows a rudimentary form of MLM (i.e., predicting a masked target word from a context window without any perturbation of the masked word).

In this work, we find that the performance discrepancy between \cbow and \sg embeddings is founded less on theoretical differences in their training objectives but more on faulty \cbow implementations in standard libraries such as the official implementation \mikolov \citep{mikolov2013distributed}, \gensim \citep{rehurek_lrec} and fastText \citep{bojanowski2017enriching}.
Specifically, we find that in these implementations, the gradient for source embeddings is incorrectly multiplied by the context window size, resulting in incorrect weight updates and inferior performance. 

We make the following contributions.  
First, we show that our correct implementation of \cbow indeed yields word embeddings that are fully competitive with \sg while being trained in less than a third of the time
(e.g., with a single machine it takes less than 73 minutes to train \cbow embeddings on 
the entire Wikipedia corpus).
We present experiments on intrinsic word similarity and analogy tasks, as well as extrinsic eval  uations on the SST-2, QNLI, and MNLI tasks from the GLUE benchmark \citep{wang2018glue}, and the CoNLL03 English named entity recognition task \citep{sang2003introduction}.

Second, we make our implementation, \us,
publicly available.\footnote{\dlurl}  
With this implementation, it is possible to train 768-dimensional \cbow embeddings on one epoch of English C4 \citep{raffel2020exploring} in 1.61 days on a single 16 CPU machine.\footnote{See Appendix~\ref{app:training_time} for training time benchmarks of popular Word2vec implementations against \us.}


\section{\cbow Implementation}
\label{sec:implementation}


The parameters of \cbow are two sets of word embeddings: 
``source-side'' and ``target-side'' vectors $v_w, v'_w \in \R^d$ for every 
word type $w \in V$ in the vocabulary. 
A window of text in a corpus consists of a center word $w_O$ 
and context words $w_1 \ldots w_C$.
For instance, in the window \texttt{the dog laughed}, 
we have $w_O = \texttt{dog}$ and $w_1 = \texttt{the}$ and $w_2 = \texttt{laughed}$. Given a window of text, the \cbow loss is defined as:
\begin{align*}
    v_c &= \frac{1}{C} \sum_{j=1}^C v_{w_j} \\
    \mathcal{L} &= - \log \sigma ( {v'_{w_O}}^\top v_{c} )  -  \sum_{i=1}^{k} \log \sigma( - {v'_{n_i}}^\top v_{c}) 
\end{align*}
where $n_1 \ldots n_k \in V$ are negative examples drawn iid from some noise distribution $P_n$ 
over $V$. 
The gradients of $\mathcal{L}$ with respect to the target ($v'_{w_O}$), negative target ($v'_{n_i}$), and average context source ($v_{c}$) embeddings are:
\begin{align}
    \frac{\partial \mathcal{L}}{\partial v'_{w_O}} = & (\sigma({v'_{w_O}}^\top v_{c}) - 1) v_{c} \notag \\
    \frac{\partial \mathcal{L}}{\partial v'_{n_i}} = & \sigma({v'_{n_i}}^\top v_{c}) v_{c} \notag \\
    \frac{\partial \mathcal{L}}{\partial v_{c}} = & (\sigma({v'_{w_O}}^\top v_c) - 1) v'_{w_O} + \notag \\ & \sum_{i=1}^{k} \sigma({v'_{n_i}}^\top v_{c}) v'_{n_i} \label{eq:incorrect}
\end{align}
and by the chain rule with respect to a source context embedding:
\begin{align}
    \frac{\partial \mathcal{L}}{\partial v_{w_j}} = \frac{1}{C} [ & (\sigma({v'_{w_O}}^\top v_c) - 1) v'_{w_O} + \notag \\
    & \sum_{i=1}^{k} \sigma({v'_{n_i}}^\top v_{c}) v'_{n_i} ] \label{eq:correct}
\end{align}
However, the \cbow negative sampling implementations in \mikolov\footnote{\url{https://github.com/tmikolov/word2vec/blob/20c129af10659f7c50e86e3be406df663beff438/word2vec.c##L483}},
\gensim\footnote{
\url{https://github.com/RaRe-Technologies/gensim/blob/a93067d2ea78916cb587552ba0fd22727c4b40ab/gensim/models/word2vec_inner.pyx##L455-L456}.},
and fastText\footnote{\url{https://github.com/facebookresearch/fastText/blob/a20c0d27cd0ee88a25ea0433b7f03038cd728459/src/model.cc##L85}
In fastText, the normalization option is guarded by a boolean flag, but it defaults to \texttt{false} for CBOW.}
 incorrectly update each context vector, $v_{w_j}$, by Eq.~\eqref{eq:incorrect}, without normalizing by the number of context words, given in Eq.~\eqref{eq:correct}.  In fact, this error has been pointed out in several \gensim issues as well as a fastText issue.\footnote{\url{https://github.com/RaRe-Technologies/gensim/issues/1873}, \url{https://github.com/RaRe-Technologies/gensim/issues/697}, \url{https://github.com/facebookresearch/fastText/issues/910}.}

\paragraph{Why This Error Matters}  Aside from being incorrect, this update matters for two reasons.  First, in both \gensim and \mikolov, the width of the context window is randomly sampled from ${1, \ldots, C_{\text{max}}}$ for every target word.  This means that source embeddings which were averaged over wider context windows will experience a larger update than their contribution, relative to source embeddings averaged over narrower windows. 
Second, in the incorrect gradient, $\frac{\partial \mathcal{L}}{\partial v}$ is incorrectly scaled by $C$, while $\frac{\partial \mathcal{L}}{\partial v'}$ is not; the correct stochastic gradient with respect to all embeddings actually points in a different direction than what was implemented in \mikolov. 
\Cref{sec:discussion} touches on both of these issues in more detail.
 
 %




\section{Experiments}
\label{sec:experiments}

We evaluated \cbow{} and \sg{} embeddings under \gensim and our implementations (both corrected and original CBOW).  Unless otherwise noted, we learn word embeddings on the entire Wikipedia corpus PTB sentence split and tokenized with default settings, where words occurring fewer than ten times were dropped.  Training hyperparameters were held fixed for each set of embeddings: negative sampling rate 5, maximum context window of 5 words, number of training epochs 5, and embedding dimensionality 300 unless otherwise noted.

We found that the default initial learning rate in \gensim, 0.025, learned strong \sg{} embeddings.  However, we swept separately for \cbow initial learning rate for \gensim and our implementation.  We swept over initial learning rate in $\{0.025, 0.05, 0.075, 0.1\}$ selecting 0.025 to learn \gensim \cbow embeddings and 0.075 for corrected \cbow, to maximize average performance on the development fold (random partition of 50\% of examples) of intrinsic evaluation tasks described in \ref{subsec:experiments:evaluation}. We selected learning rate as this was a critical hyperparameter, and we found that \cbow embeddings learned with 0.075 learning rate with \gensim suffered compared to learning with a low learning rate.  

\subsection{Evaluation}
\label{subsec:experiments:evaluation}

We evaluated each set of embeddings intrinsically on the \texttt{MEN}, \texttt{WS353}, \texttt{MIXED}, and \texttt{SYNT} word similarity and analogy tasks, as described in \citet{levy2014neural}.  We also evaluate on the Stanford rare words analogy task, \texttt{RW} \citep{luong2013better}.  Analogy tasks were evaluated by top-1 accuracy and similarity tasks using Spearman's rank correlation coefficient.

We also evaluated embeddings extrinsically on the SST-2, QNLI, and MNLI GLUE tasks, following the methodology in \citet{wang2018glue} and use the frozen pre-trained word embeddings in a BiLSTM classifier.  In addition, we also evaluated word embeddings in a CoNLL03 English named entity recognition (NER) sequence tagger \cite{sang2003introduction}.  For NER, we use a single-layer BiLSTM with 256-dimensional hidden layer as the sequence tagger, and evaluate under frozen and finetuned word embedding settings.  See Appendix~\ref{app:hyperparameters} for model selection details in extrinsic evaluation.

\section{Results}
\label{sec:results}

\paragraph{Intrinsic Evaluation} \cbow embeddings trained with our implementation achieve similar intrinsic performance to \sg embeddings, surpassing \sg embeddings for the \texttt{RW} and \texttt{SYNT} tasks (\Cref{tab:intrinsic_eval}).  On the other hand, \cbow embeddings trained by \gensim or our incorrect implementation achieve much worse performance than \sg.  We also report performance of \cbow embeddings, with and without the corrected update, trained for one epoch on the entirety of the English C4 dataset (768 dimensions and 5 million word vocabulary).  These embeddings took 1.61 days per epoch to train on a single 16-core machine.

\begin{table*}[!ht]
\centering
\footnotesize
\begin{tabular}{l|l|c|c|c|c|c|c} \hline
{\bf Package} & {\bf Objective} & {\bf ws353} & {\bf men} & {\bf rw} & {\bf syn}\ & {\bf mixed} & {\bf AVG} \\ \hline
\multirow{2}{*}{\textbf{\gensim}} & \sg & 72.2/61.9 & 72.1/72.5 & 46.8/44.9 & 72.8/72.4 & 82.5/81.3 & 69.3/66.6 \\
 & \cbow & 61.5/62.6 & 69.2/70.0 &  43.2/38.1 &  69.9/71.1 & 79.1/78.0 & 64.6/64.0 \\ \hline
\multirow{5}{*}{\textbf{\us}} & \sg & 72.8/62.0 & 72.7/72.0 & 44.9/43.8 & 73.6/71.8 & 83.0/80.2 & 69.4/66.0 \\
 & \cbow [s]& 61.2/62.6 & 68.8/69.4 & 48.6/39.9 & 73.4/74.7 & 78.3/76.8 & 66.1/64.7 \\  
 & \cbow [f] & 70.6/65.8 & 74.0/74.6 & 49.0/45.5 & 76.7/76.5 & \textbf{83.8/82.1} & 70.8/68.9 \\
  & \cbow [s]; C4 & 72.9/68.0 & 81.0/79.9 & 51.1/50.3 & 80.5/79.9 & 81.5/81.1 & 73.4/71.8 \\
  & \cbow [f]; C4 & \textbf{74.1/68.1} & \textbf{79.8/80.2} & \textbf{53.1/54.8} & \textbf{82.9/81.6} & 82.2/81.9 & \textbf{74.4/73.3} \\ \hline
     \end{tabular}
\caption{Intrinsic evaluation of Wikipedia-trained Word2vec embeddings on \texttt{dev/test} folds.  Spearman's rank correlation coefficient is reported for: \texttt{wordsim353}, \texttt{men}, and \texttt{rw}, and accuracy for: \texttt{syn} and \texttt{mixed}.  \texttt{AVG} is the average across all five tasks. The best test performance for each task is bolded. \emph{CBOW [s]} refers to the standard, incorrect implementation of CBOW, and \emph{CBOW [f]} is the fixed version.}
\label{tab:intrinsic_eval}
\end{table*}

\paragraph{Extrinsic Evaluation}
Development and test set performance on GLUE tasks for each set of embeddings is reported in \Cref{tab:extrinsic_glue_eval}.  We also compare performance against 300-dimensional GloVe embeddings \citep{pennington2014glove} pre-trained on Wikipedia and Gigaword 5 and random embeddings drawn from a standard normal distribution as baselines.

\begin{table*}[!ht]
\centering
\small
\begin{tabular}{l|l|c|c|c|c} \hline
{\bf Package} & {\bf Objective} & {\bf SST-2} & {\bf QNLI} & {\bf MNLI-m} & {\bf MNLI-mm} \\ \hline
\multirow{2}{*}{\textbf{Baselines}} & Random & 83.49/81.4 & 63.88/63.9 & 61.41/60.3 & 61.41/59.6\\
& GloVe & 85.09/83.6 & 66.37/67.7 & 66.93/66.7 & 66.93/66.1\\ \hline
\multirow{2}{*}{\textbf{\gensim}} & \sg & 86.93/85.6 & 69.39/67.9 & 68.02/67.3 & 68.02/67.9 \\
& \cbow & \textbf{88.07/86.7} & \textbf{69.06/70.7} & 68.15/68.2 & 68.15/66.9 \\ \hline
\multirow{2}{*}{\textbf{\us}} & \sg & 86.24/84.5 & 69.06/68.2 & 68.20/67.4 & 68.20/67.7 \\
 & \cbow & 88.42/85.3 & 68.17/68.6 & \textbf{69.22/68.4} & \textbf{69.22/68.6} \\ \hline
    \end{tabular}
\caption{Percent accuracy of word embeddings on the dev/test sets of the SST, QNLI, and MNLI GLUE tasks.
In the case of MNLI, development accuracy is averaged over the matched and mismatched development
sets, which is also used for model selection.  Test performance is reported separately for matched and mismatched domains. The best test performance for each task is in bold.}
\label{tab:extrinsic_glue_eval}
\end{table*}

\Cref{tab:extrinsic_ner_eval} contains CoNLL 2003 NER performance for each set of word embeddings.  We used the same set of hyperparameters sampled in the GLUE evaluation for model selection (dev F1 as computed by the CoNLL evaluation script).  NER test performance of corrected \cbow embeddings is within 1\% F1 to \sg embeddings, whereas \gensim \cbow embeddings suffer up to 4\% F1 test performance relative to \sg.

\begin{table}[!ht]
\centering
\footnotesize
\begin{tabular}{l|l|p{0.6cm}p{0.6cm}|p{0.6cm}p{0.6cm}} \hline
 & & \multicolumn{2}{c}{\bf Frozen} & \multicolumn{2}{c}{\bf Finetuned} \\ \hline
{\bf Package} & {\bf Objective} & {\bf Dev} & {\bf Test} & {\bf Dev} & {\bf Test} \\ \hline
\multirow{3}{*}{\textbf{\gensim}}  & Random & 83.6 & 75.4 & 83.0 & 74.3 \\
& \sg & 92.5 & {\bf 88.4} & 92.9 & {\bf 88.2} \\
& \cbow & 90.1 & 84.4 & 90.0 & 85.3 \\ \hline
\multirow{2}{*}{\textbf{\us}} & \sg & 92.5 & 88.2 & 92.9 & 88.0 \\
 & \cbow & 92.3 & 87.8 & 92.3 & 87.4 \\ \hline
    \end{tabular}
\caption{Percent F1 on the CoNLL 2003 English NER heldout sets for each set of word embeddings.}
\label{tab:extrinsic_ner_eval}
\end{table}

\section{Discussion}
\label{sec:discussion}

The \gensim-trained \cbow embeddings achieve better test accuracy on SST-2 despite achieving similar performance on the development set as corrected \cbow embeddings. \gensim \cbow embeddings also outperform on QNLI.  However, performance on MultiNLI is slightly different.  In this case, corrected \cbow embeddings are almost 2\% more accurate than \gensim embeddings when evaluated on the out of domain test set. For these GLUE tasks, it is unclear whether \cbow embeddings learned with our implementation are universally more performant than those learned by \gensim.  We hypothesize this may be due to the susceptibility of models overfitting to the development set (especially since our classifier has 1024 dimensional hidden layers).

\subsection{Problems with Faulty \cbow} 
\label{subsec:implementation:faulty_implementation_problems}

Aside from being incorrect, the faulty update leads to two major issues during training: (1) the norm of the source side vectors grows as a function of context width, and (2) despite being a descent direction, the faulty update diverges from the SGD direction as the number of negative samples increases.

\paragraph{Embedding Norm Grows with Context}

Because the effective learning rate for source embeddings is greater than that for the source embeddings, one would expect the magnitude of the source embeddings to also grow larger.  
Growing source norms can be a problem during learning, as larger embedding norms can easily lead to saturated sigmoid activations, effectively halting learning. See Appendix~\ref{app:norm} for an empirical analysis of embedding magnitude.

\paragraph{Update Worsens with More Negatives}

The faulty \cbow update for source embeddings is a stochastic\footnote{Context width is sampled uniformly at random between $1$ and maximum context width in standard word2vec implementations.} multiple of the correct update.  However, \cbow learns both source and target embeddings, which means that the update direction no longer follows the SGD update direction.

Even though the faulty update is still a descent direction, we can analyze how the angle between it and the true SGD direction changes as a function of context width. If we make the simplifying assumption that the norm of the source and target gradients are equal, then one can derive the cosine similarity between $\delta \theta$ (true gradient) and $\widetilde{\delta \theta}$ (faulty gradient) with respect to \cbow parameters $\theta$ as:

\begin{align}
    \text{cos}(\widetilde{\delta \theta}, \delta \theta) = & \frac{C^{2} + k + 1}{\sqrt{(C^{3} + k + 1) (C + k + 1)} }
    \label{eq:cosim_grad}
\end{align}

\begin{figure}
    \centering
    \includegraphics[clip, trim=1.5cm 0.75cm 1.8cm 1.5cm, width=0.9\linewidth]{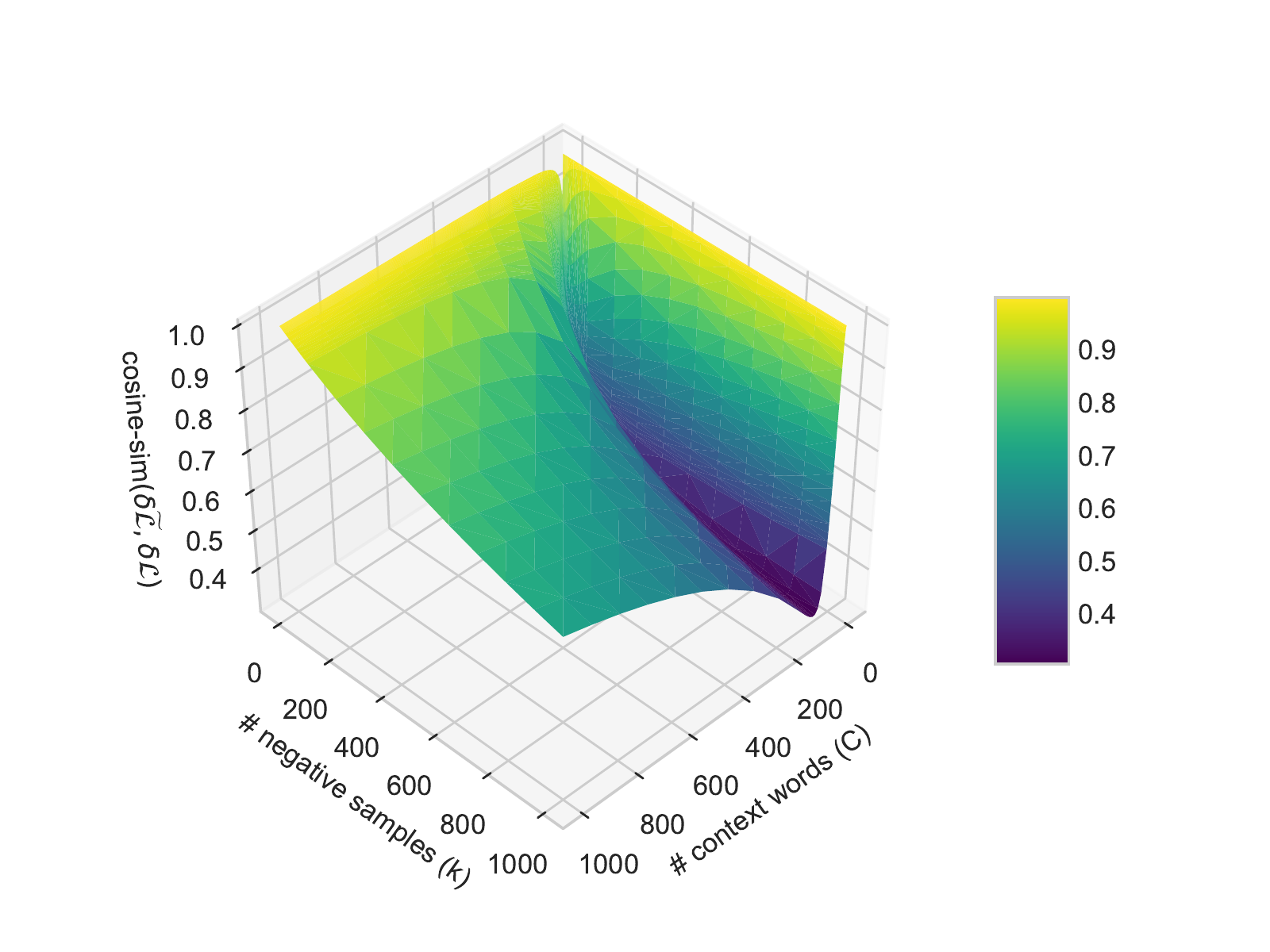}
    \caption{Cosine similarity of incorrect to correct stochastic gradient as a function of $C$ and $k$ assuming that the norm of the gradient with respect to each embedding is fixed to a shared constant.  The minimum cosine similarity displayed in this plot is 0.303.}
    \label{fig:cosim_cbow_grad}
\end{figure}

Where $C$ is the number of context tokens and $k$ is the number of negative samples. 
For a moderate context width, as the number of negative examples increases, the faulty update points further away from the true gradient (\Cref{fig:cosim_cbow_grad}).  See Appendix~\ref{app:analysis} for the full derivation.

\begin{figure}
    \centering
    \includegraphics[width=0.9\linewidth]{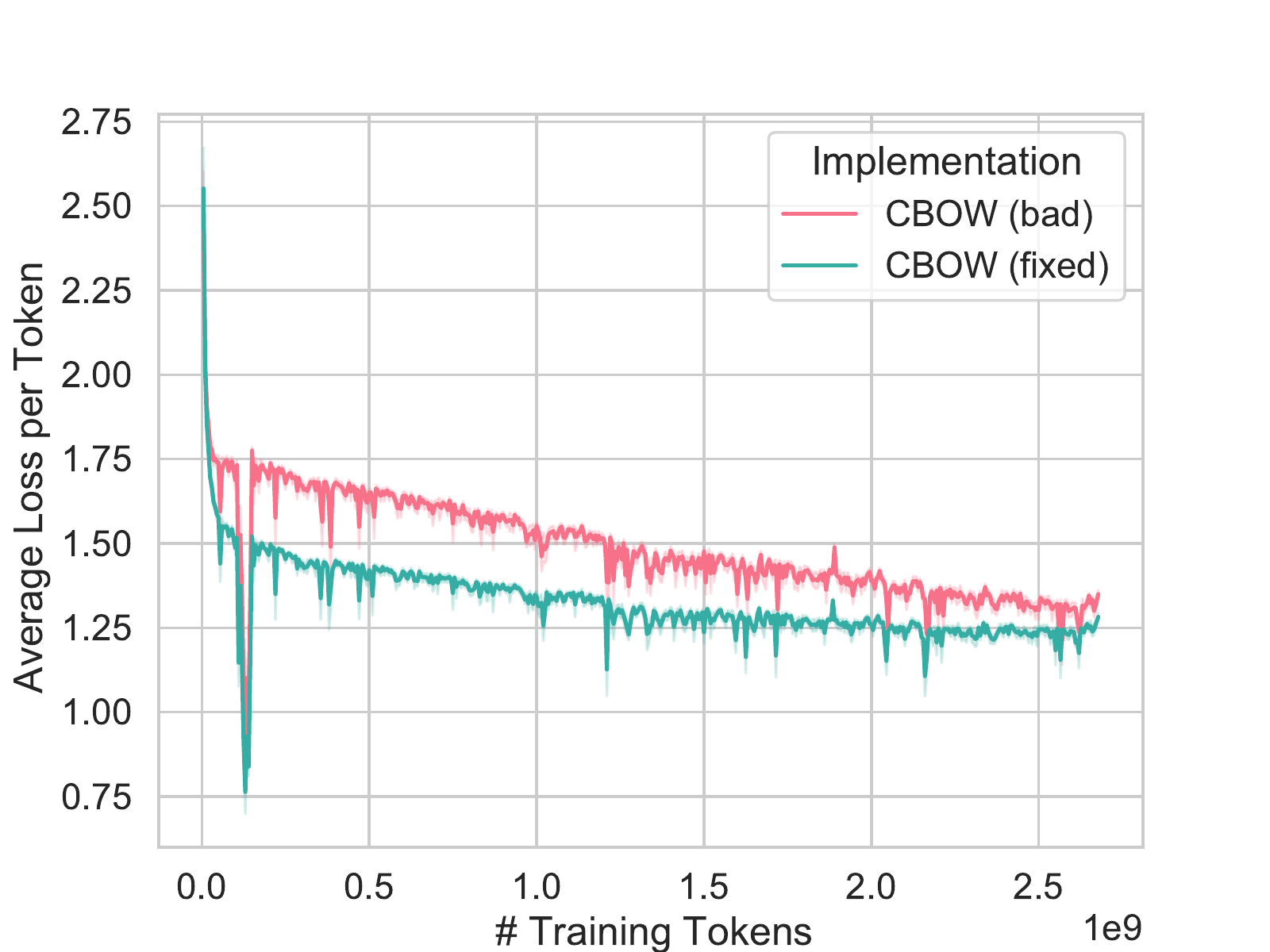}
    \caption{Average negative sampling loss per token for every batch of 5 million tokens for a single epoch of CBOW training on Wikipedia.  The shaded region corresponds to the 95\% bootstrapped confidence interval over average token loss on 100K token batches.}
    \label{fig:training_curves}
\end{figure}


In practice, we found performance of \cbow embeddings to only decay moderately as a function of number of negatives, regardless of correction to the update, consistent with that reported in \citet{pennington2014glove}.  Nevertheless, we find that our correct \cbow implementation decreases training loss more quickly than the incorrect update, even in the typical setting of 5 negative samples and context width of 5 (\Cref{fig:training_curves}).  Ultimately, increased sensitivity to hyperparameters is not a concern under typical training regimes.

\paragraph{FastText} Although we do not investigate fastText specifically in this work, we do note in \Cref{sec:implementation} that fastText \cbow defaults to using the same incorrect source embedding gradient. Sent2vec \cite{gupta-etal-2019-better} is also exposed to this bug, since it builds on fastText. In spite of this, fastText embeddings trained with the \cbow objective were found to outperform \sg across multiple languages \cite{grave2018learning}. To achieve better performance of \cbow, \citet{grave2018learning} tuned hyperparameters and trained on a web-scale corpus (enabled by the faster \cbow training).

FastText also represents each word vector as the sum of vectors of its constituent character n-grams, and includes positional embeddings as part of the training procedure. In this work, we only consider the original \cbow and \sg models described in \citet{mikolov2013efficient} and hold hyperparameters and training set fixed between models whenever possible. We leave investigation of whether correcting this gradient bug could further improve fastText embeddings as future work.



\section{Conclusion}
\label{sec:conclusion}


Before the widespread adoption of automatic differentiation libraries, it was the modeler's responsibility to derive the correct gradient for updating model weights. This \mikolov gradient bug could have been avoided if a finite difference check was run to ensure the derivation of the stochastic gradient was correct. These checks are standard practice \cite{bengio2012practical,bottou2012stochastic}, and we strongly encourage other researchers to use finite difference gradient checks to verify the correctness of manually derived gradients. According to \citet{bottou2012stochastic}: ``\emph{When the computation of the gradients is slightly incorrect, stochastic gradient
descent often works slowly and erratically... It is not
uncommon to discover such bugs in SGD code that has been quietly used for years.}''

We find that \cbow can learn embeddings that are as performant as \sg, when trained with the correct update, allowing efficient training of strong word embeddings on web-scale datasets on an academic budget.  We release our implementation, \us, along with trained C4 \cbow embeddings at \dlurl.

\bibliography{insights_ws2021/main}
\bibliographystyle{insights_ws2021/acl_natbib}

\clearpage

\appendix

\appendix 

\section{Versions of \gensim and \mikolov}
\label{app:versions}

We compare against the \gensim implementation\footnote{\url{https://github.com/RaRe-Technologies/gensim/blob/develop/gensim/models/word2vec.py}}
at commit 

{
\footnotesize
\texttt{a93067d2ea78916cb587552ba0fd22727c4b40ab}
}

and the \mikolov implementation\footnote{\url{https://github.com/tmikolov/word2vec/}} at commit 

{
\footnotesize
\texttt{e092540633572b883e25b367938b0cca2cf3c0e7}
}

which are the most recent commits at the time of writing.

\section{Training time}
\label{app:training_time}

\begin{figure}
    \centering
    \includegraphics[width=0.45\linewidth,page=1]{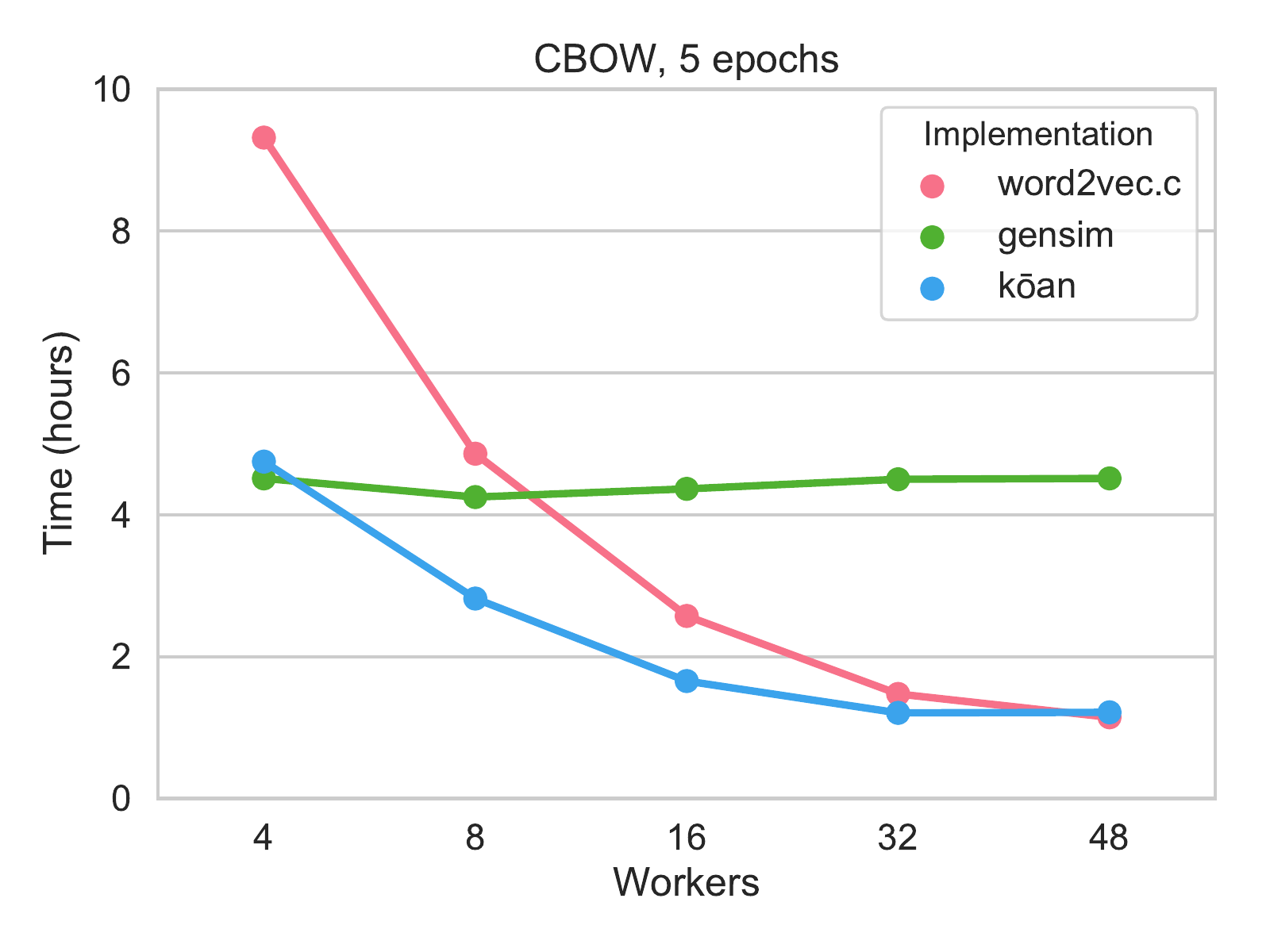}
    \includegraphics[width=0.45\linewidth,page=2]{figures/word2vec_train_times.pdf}
    \caption{Hours to train \cbow{} (left) and \sg{} (right) embeddings as a function of number of worker threads for different Word2vec implementations.}
    \label{fig:training_time}
\end{figure}


\Cref{fig:training_time} shows the time to train each set of embeddings as a function of number of worker threads on the tokenized Wikipedia corpus.  All implementations were benchmarked on an Amazon EC2 c5.metal instance (96 logical processors, 196GB memory) using \texttt{time}, and embeddings were trained with a minimum token frequency of 10 and downsampling frequency threshold of $10^{-3}$. We benchmark our implementation with a buffer of 100,000 sentences.  We trained \cbow{} for five epochs and \sg{} for a single epoch, as it is much slower.  \gensim{} seems to have trouble exploiting more worker threads when training \cbow, and both \mikolov and our implementation learn embeddings faster.
\mikolov achieves slightly better scaling than our implementation during \cbow{} training (68m46s vs. 72m57s for 48 threads), although ours is faster with fewer workers (e.g., 559m0s vs. 284m58s for 4 threads and 88m28s vs. 72m29s for 32 threads). Although we did not comprehensively benchmark \sg{} training for 5 epochs, we found that training \sg{} embeddings for 5 epochs with 48 threads took us 3.31 times as long as training \cbow.

\section{Extrinsic Evaluation Hyperparameters}
\label{app:hyperparameters}

We evaluate a given set of embeddings on GLUE tasks by using the frozen pre-trained word embeddings in a neural classifier.  Words that were out of vocabulary are represented with a zero embedding vector.   We use a two layer BiLSTM with 1024 hidden layer width, and a final projection layer.  For the NLI tasks, we encode each sequence with an independent BiLSTM encoder.  If the embeddings of sequence 1 and 2 are $v$ and $w$, respectively, then a prediction is made using a multilayer perceptron over $[v ; w ; v - w ; v \odot w]$ with a single 512 unit hidden layer.  This is identical to the NLI classifier described in \citet{wang2018glue}. We performed a random search over dropout rate $\in [0.0, 0.7]$ and learning rate $\in [10^{-6}, 10^{-2}]$ with a budget of ten runs.  All models are trained for up to 100 epochs with a patience of 2 epochs for early stopping.  The same set of sampled hyperparameters was used for model selection for each set of word embeddings.

\section{\cbow Embedding Norm}
\label{app:norm}

\begin{figure}
    \centering
    \includegraphics[width=0.9\linewidth]{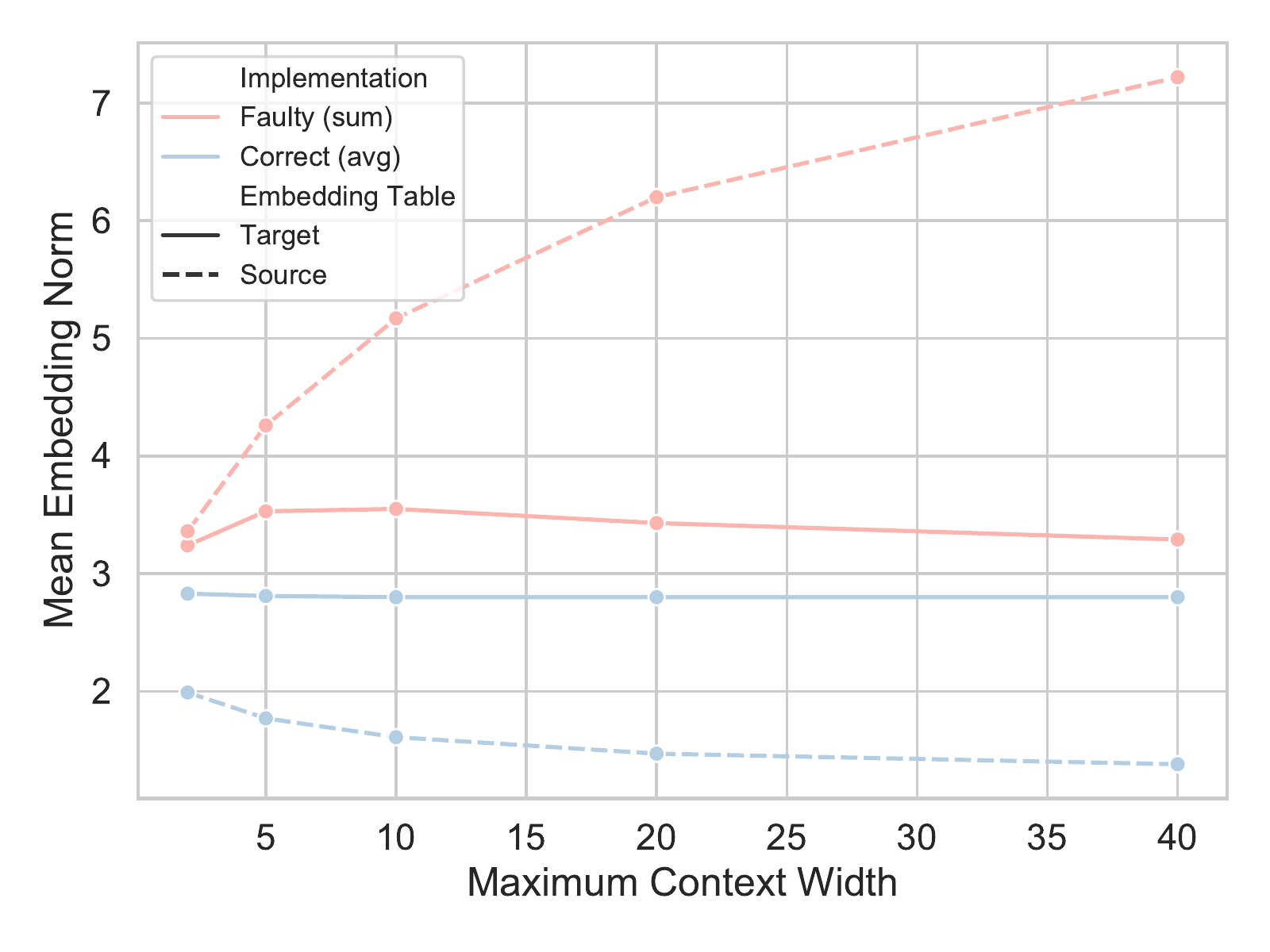}
    \caption{Average target and source vector norms as a function of context width for the correct and incorrect implementations of \cbow.  All embeddings were trained for five epochs over a one million sentence sample of the Wikipedia corpus.}
    \label{fig:emb_norm}
\end{figure}

If we train embeddings using the correct implementation of \cbow, the $l_2$ norms of target and source vectors are unaffected by the width of the context window
(\Cref{fig:emb_norm}).
With the faulty implementation \Cref{eq:incorrect}, the source vector norms increase as a function of the maximum context window width.  Growing source norms can be a problem during learning, as larger embedding norms can easily lead to saturated sigmoid activations, effectively halting learning.  This problem is further exacerbated by fast implementations of \cbow{} and \sg{} that approximate the sigmoid activation function by a piecewise linear function.  In these approximations, when the logit is above or below some threshold (e.g., 6 and -6 for \mikolov) and the prediction agrees with ground truth, the gradient for this example is not back-propagated at all.

\REMOVED{
\begin{table*}[!ht]
\centering
\footnotesize
\begin{tabular}{c|c|c|c|c} \hline
& \multicolumn{2}{c}{{\bf Correct (avg)}} & \multicolumn{2}{c}{{\bf Incorrect (sum)}} \\
{\bf Context width} & {\bf Mean target norm} & {\bf Mean source norm} & {\bf Mean target norm} & {\bf Mean source norm} \\ \hline
2 & 2.83 & 1.99 & 3.24 & 3.36 \\
5 & 2.81 & 1.77 & 3.53 & 4.26 \\
10 & 2.80 & 1.61 & 3.55 & 5.17 \\
20 & 2.80 & 1.47 & 3.43 & 6.20 \\
40 & 2.80 & 1.38 & 3.29 & 7.22 \\ \hline
    \end{tabular}
\caption{Average target and source vector norms as a function of context width for the correct and incorrect implementations of \cbow.  All embeddings were trained for five epochs over a one million sentence sample of the Wikipedia corpus.}
\label{tab:emb_norm}
\end{table*}
}

\section{Analysis of the Incorrect \cbow Gradient}
\label{app:analysis}

\paragraph{Relationship to correct \cbow{} update} Although the incorrect \cbow{} update for source and target embeddings points in a different direction than the SGD update, this is still a descent direction for the \cbow{} loss.  In other words, taking a sufficiently small step in this direction will reduce the stochastic loss.  This can be easily seen by:

\begin{align}
  D = & \left(\begin{array}{c|c}
              C I_{d C} & 0 \\ \hline
              0 & I_{d (k+1)}\end{array}\right) \nonumber \\
  \delta v = & [\frac{\partial \mathcal{L}}{\partial v_{w_1}} ; \ldots ; \frac{\partial \mathcal{L}}{\partial v_{w_C}} ] \nonumber \\
  \delta v' = & [ \frac{\partial \mathcal{L}}{\partial v'_{n_1}}; \ldots ; \frac{\partial \mathcal{L}}{\partial v'_{n_k}} ; \frac{\partial \mathcal{L}}{\partial v'_{w_O}} ] \nonumber \\
  \delta \theta = & [ \frac{\partial \mathcal{L}}{\partial v} ; \frac{\partial \mathcal{L}}{\partial v'} ] \nonumber \\
  \widetilde{\delta \theta} = & D \left( \delta \theta \right)
\end{align}

where $d$ is the word embedding dimensionality, $D$ is a diagonal matrix that scales the gradient with respect to each context word by $C$, and leaves the gradient with respect to the negative sampled target words and true target word unchanged.  $\delta \theta$ and $\widetilde{\delta \theta}$ denote the correct and incorrect gradients of the loss, respectively, with respect to source and target embeddings.  Since all diagonal entries on $D$ are strictly positive, $D$ is positive definite.  Therefore, if $\|\delta \theta\| > 0$ then $(- \widetilde{\delta \theta})^T \delta \theta < 0$ and $- \widetilde{\delta \theta}$ is a descent direction for the stochastic loss \citep{boyd2004convex}.

Even though the incorrect negative gradient is guaranteed to be a descent direction, the angle between this descent direction and the negative gradient can be influenced by the number of negative samples and the number of context words to average over.  For the sake of simplicity, suppose that the gradient of the loss with respect to each source and target embedding has the same norm: $\forall j \in \{1, \ldots, C\}, i \in \{1, \ldots, k\}: \|\frac{\partial L}{\partial v_{w_j}}\|^2_2 = \|\frac{\partial L}{\partial v'_{n_i}}\|^2_2 = \alpha$.  Then the cosine similarity between the incorrect and correct gradient can be written as:

\begin{align}
    \texttt{cos}(\widetilde{\delta \theta}, \delta \theta) & = \frac{(\widetilde{\delta \theta})^T \delta \theta}{\| \widetilde{\delta \theta} \|_2 \| \delta \theta \|_2} & \notag \\
    & = \frac{(C^2 + k + 1) \alpha}{\sqrt{(C^3 + k + 1) \alpha} \sqrt{(C + k + 1) \alpha} } \notag \\
    & = \frac{ (C^2 + k + 1) }{\sqrt{(C^{3} + k + 1) (C + k + 1)}} \notag
\end{align}

\REMOVED{
\begin{figure}
    \centering
    \includegraphics[width=0.9\linewidth]{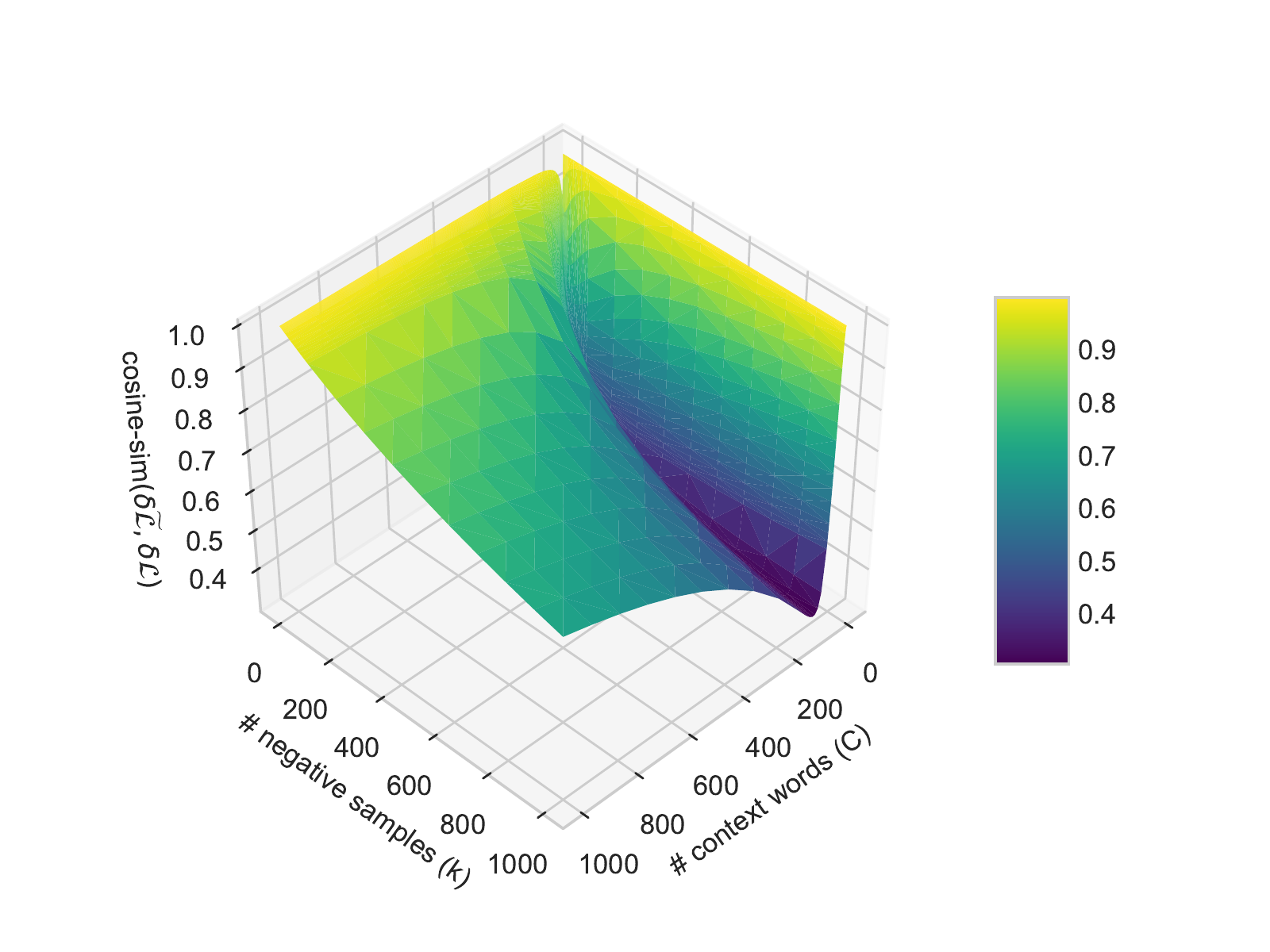}
    \caption{Cosine similarity of incorrect to correct stochastic gradient as a function of $C$ and $k$ assuming that the norm of the gradient with respect to each embedding is fixed to a shared constant.  The minimum cosine similarity displayed in this plot is 0.303.}
    \label{fig:cosim_cbow_grad}
\end{figure}
}

We can better understand how the angle between the incorrect and correct gradient varies as a function of number of context words and negative samples by looking at a plot of this function (\Cref{fig:cosim_cbow_grad}).  What this plot makes clear is that for a moderate number of context words (around 10, which corresponds to a maximum context window of 5), the incorrect stochastic gradient can differ significantly from the true gradient as the number of negative samples increases.  However, for sensible settings of $C, k \leq 20$, the minimum cosine similarity is 0.68, achieved by $C=9$ and $k=20$, with 0.82 cosine similarity with $C=5$ and $k=5$ (typical for Word2vec training).  Because of this, the \cbow{} update bug may have gone unnoticed.  If one were to sample a large number of negatives, then the bug may have been more apparent.

\end{document}